\documentclass[lettersize,journal]{IEEEtran}
\usepackage{amsmath,amsfonts}
\usepackage{algorithmic}
\usepackage{array}
\usepackage[caption=false,font=normalsize,labelfont=sf,textfont=sf]{subfig}
\usepackage{textcomp}
\usepackage{stfloats}
\usepackage{hyperref}
\usepackage{url}
\usepackage{verbatim}
\usepackage{graphicx}
\usepackage{arydshln}
\usepackage{booktabs}
\usepackage{multirow}
\usepackage{soul}
\usepackage{xcolor}

\def\BibTeX{{\rm B\kern-.05em{\sc i\kern-.025em b}\kern-.08em
    T\kern-.1667em\lower.7ex\hbox{E}\kern-.125emX}}
\usepackage{balance}
\begin{document}
\title{REGEN: Real-Time Photorealism Enhancement in Games via a Dual-Stage Generative Network Framework}
\author{Stefanos Pasios, Nikos Nikolaidis
\thanks{The authors are with the School of Informatics, Aristotle University of Thessaloniki, Thessaloniki 54124, Greece (email: \{pstefanos,nnik\}@csd.auth.gr).}}

\maketitle

\begin{abstract}
Photorealism is an important aspect of modern video games since it can shape player experience and impact immersion, narrative engagement, and visual fidelity. To achieve photorealism, beyond traditional rendering pipelines, generative models have been increasingly adopted as an effective approach for bridging the gap between the visual realism of synthetic and real worlds. However, under real-time constraints of video games, existing generative approaches continue to face a tradeoff between visual quality and runtime efficiency. In this work, we present a framework for enhancing the photorealism of rendered game frames using generative networks. We propose REGEN, which first employs a robust unpaired image-to-image translation model to generate semantically consistent photorealistic frames. These generated frames are then used to create a paired dataset, which transforms the problem to a simpler unpaired image-to-image translation. This enables training with a lightweight method, achieving real-time inference without compromising visual quality. We evaluate REGEN on Unreal Engine, showing, by employing the CMMD metric, that it achieves comparable or slightly improved visual quality compared to the robust method, while improving the frame rate by 12×. Additional experiments also validate that REGEN adheres to the semantic preservation of the initial robust image-to-image translation method and maintains temporal consistency. Code, pre-trained models, and demos for this work are available at: \url{https://github.com/stefanos50/REGEN}

\end{abstract}


\section{Introduction}
\IEEEPARstart{P}{hotorealism} in modern video games is a key factor in shaping player experience. High-quality graphics not only enhance the immersion and realism but also support the narrative, emotional investment, and understanding of the game. In competitive markets where first impressions of the target audience are critical, photorealistic visuals alone can make or break a video game's success. Despite considerable advances in the computation capabilities of modern hardware and the development of sophisticated computer graphics pipelines with state-of-the-art technologies such as real-time ray tracing and lumen, achieving true photorealism in dynamic and interactive environments continues to be a significant challenge. Conventional rendering methods often have to compromise on quality to achieve sufficient performance, particularly in real-time conditions where a constant, high enough frame rate is of utmost importance.

To overcome the limitations of traditional computer graphics pipelines, recent advances in Artificial Intelligence (AI), particularly in generative models \cite{Sengar2025}, have opened new opportunities for bridging the gap between games and reality. Image-to-Image (Im2Im) translation \cite{s22218540} is an approach that is primarily utilized to enhance the photorealism of real-time rendered images based on a target domain of the real world \cite{Richter_2021, pasios2024carla2realtoolreducingsim2real, brehm2021semanticallyconsistentimagetoimagetranslation, nvidia2025cosmostransfer1conditionalworldgeneration}. Particularly, Im2Im translation methods are more suitable for photorealism enhancement compared to other approaches (e.g., diffusion models \cite{nvidia2025cosmostransfer1conditionalworldgeneration}) since they require less computational resources, are less prone to visual hallucinations, and do not require constant prompt optimization. Im2Im translation can be either paired \cite{cyclegan, CUT}, where models learn from corresponding image pairs, or unpaired \cite{safayani2025unpairedimagetoimagetranslationcontent, wang2018pix2pixHD, pixpixold}, where mappings are learned between domains without pixel-level alignment. Since acquiring real-world images that exhibit per-pixel correspondence to the content depicted in the respective rendered game images is hard, this process is performed in an unpaired manner in the case of photorealism enhancement.

The distribution differences in unpaired Im2Im translation between the source (game) and the target (real-world) domains often introduce significant artifacts (i.e., semantic or temporal inconsistencies) that undermine player experience \cite{Richter_2021}. This has resulted in research on more robust unpaired Im2Im translation methods that utilize intermediate information \cite{Richter_2021, 10458434}. Such information is employed during image rendering and includes various properties of the virtual environment, such as depth, geometry, and materials. However, these methods have not been integrated into real-time rendering pipelines, since they require access to information that is integrated deep into the engine and run at less than 10 frames per second (FPS).

In this paper, we propose Real-time photorealism Enhancement in Games via a dual-stage gEnerative Network framework (REGEN), a new framework to perform photorealism enhancement in games that is easy to integrate and runs in real-time. In detail, we propose the utilization of a robust unpaired Im2Im translation method \cite{Richter_2021} that can generate highly photorealistic and semantically consistent results in non-real-time. Considering that the generated data closely aligns semantically with the initial rendered frames of the game, the task is transformed into a paired Im2Im translation \cite{wang2018pix2pixHD, RottShaham2020ASAP} problem, which is a significantly easier process. Paired training provides direct supervision for each input-output pair, removing the ambiguity present in unpaired translation and allowing the model to learn precise mappings with fewer artifacts; as a result, the task can be performed effectively using smaller models that can be deployed in real time. Particularly, we attempt to enhance the photorealism of environments rendered with Unreal Engine (UE). Through experimentation, we illustrate both qualitatively and quantitatively that the proposed approach performance of the robust unpaired Im2Im translation method in both within- and cross-dataset evaluation and increases the FPS by a factor of 12 without model compression or additional optimizations. It is also shown that the approach adheres to the semantic preservation of the robust unpaired Im2Im translation method and maintains temporal consistency.

Our contributions are summarized as follows:

\begin{enumerate}

\item We propose a two-stage framework for photorealism enhancement in games that employs a robust unpaired image-to-image translation model to generate photorealistic pairs offline, which are subsequently used to train a lightweight paired model for real-time inference.

\item The introduced framework can be seamlessly integrated into games without requiring access to low-level engine-specific information, which is typically required as input by the current state-of-the-art (SotA) Im2Im translation methods for computer graphics.

\item Through qualitative and quantitative evaluation, we show that our approach slightly improves the visual quality of the initial unpaired robust model while providing a 12× speedup in terms of FPS, preserving semantic robustness, and maintaining temporal consistency.

\end{enumerate}

\section{Related Work} \label{prev_work}

Enhancing the photorealism of computer graphics rendered images has been widely studied in the literature, as it is a fundamental step toward improving the photorealism of visual synthetic datasets. In detail, the employment of computer-graphics-based simulators \cite{dosovitskiy2017carlaopenurbandriving} to generate large-scale synthetic datasets for training deep-learning methods has given rise to the need for enhanced photorealism. GTA V is a game that has been extensively used in deep learning research to extract synthetic datasets \cite{Richter_2016_ECCV, scucchia2025gamingresearchgtav} through various tools \cite{kiefer2021leveragingsyntheticdataobject} that transform GTA V into a deep learning data generator. Thus, several works have attempted to fill the simulation-to-reality (sim2real) gap \cite{Richter_2021, pasios2024carla2realtoolreducingsim2real, PasiosNikolaidis2025_EUSIPCO, pscsi} that exists in rendered frames and can affect the real-world performance of the models that are to be trained on those datasets.

Traditional unpaired Im2Im translation methods \cite{cyclegan, CUT, dclgan} have been extensively utilized in both simulators and games such as GTA V and have been proven to generate a significant number of artifacts \cite{Richter_2021}. The real-world datasets \cite{cityscapes, kitti} that are typically employed as the target domain for photorealism enhancement via unpaired Im2Im translation are challenging due to the different distribution between the game images and their frames. To illustrate, having a real-world dataset captured in a geographic domain with trees as the most dominant object at the top of the frames \cite{cityscapes} has been proven to result in the generation of non-realistic objects, such as trees in the sky, on the photorealism-enhanced images \cite{Richter_2021}. 

To overcome the issues of traditional unpaired Im2Im translation methods, recent works started to inject additional information \cite{brehm2021semanticallyconsistentimagetoimagetranslation, 9644765}, namely semantic segmentation information, to guide the translation process into more robust results that preserve the semantic content of the initial rendered image. Most recent advances in Im2Im translation, as well as diffusion models, for computer graphics started to employ additional information generated by the game engine to achieve robustness in various Im2Im translation tasks, including photorealism enhancement \cite{Richter_2021, nvidia2025cosmostransfer1conditionalworldgeneration} and style transfer \cite{10458434} in games. This information, typically referred to as G-Buffers, is generated through deferred rendering and provides information about scene depth, geometry (e.g., normals), and material properties (e.g., metallic, base color, and roughness) of the virtual objects. As a result, these approaches have illustrated results that can preserve the overall semantic structure and content of the scene in addition to the material properties, such as color, of the objects. Moreover, considering that most of the G-Buffers (e.g., base color, metallic, and specular) are view and sequence-independent, these methods can maintain temporal consistency, which is important in time-sequential data such as videos, games, and simulations.

Despite significant research in both paired and unpaired Im2Im translation for deep learning research and games, to date, none of the approaches has been integrated into real-time rendering pipelines for research (e.g., simulators) or video games. These methods fail to meet the dual requirement of semantic and temporal consistency in real-time environments and are typically categorized into two categories: the ones that can achieve fast inference time but are more prone to visual errors, and those that can achieve robust visual results but cannot reach real-time performance.

\begin{figure*}[htbp]
    \centering
    \includegraphics[width=0.84\textwidth]{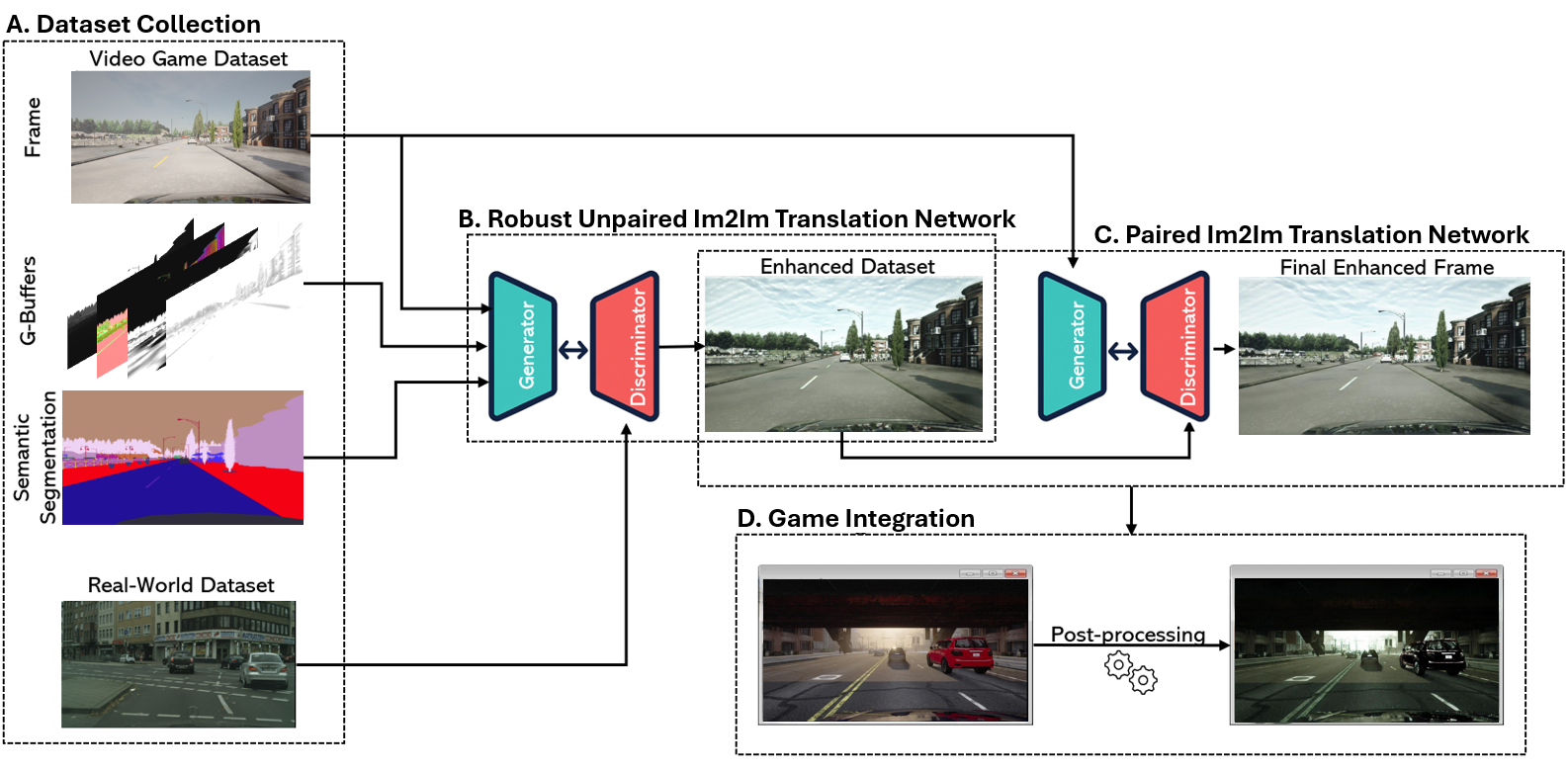}
    \caption{Overview of the REGEN framework, divided into four main phases: a) collection of an unpaired set that includes video game and real-world data, b) training of a robust unpaired Im2Im translation network to produce a semantically consistent photorealism-enhanced version of the game dataset, c) training of a lightweight paired Im2Im translation method between the video game and the photorealism-enhanced datasets, and d) final integration in the game.}
    \label{fig:flowchart}
\end{figure*}

\section{Methodology} \label{methodology}

In this section, we present the proposed framework, which can be split into four phases as illustrated in Fig. \ref{fig:flowchart}: A) collection of the video game and real-world datasets, B) training of a robust unpaired Im2Im translation method to generate the photorealism enhanced pairs of the video game dataset, C) training of a lightweight paired Im2Im translation method on the video game and the generated photorealism enhanced dataset, and D) integration of the resulting paired Im2Im translation model into the game engine.

\subsection{Dataset Collection}

In order to train a model to enhance the photorealism of video game frames, two types of datasets are required: the source dataset captured through the game and the real-world dataset. Considering that most of the robust methods utilize information that is generated through the game engine (G-Buffers), this can be a challenging task for closed-source games. An approach that can be used in this case is to detect and export this information directly from the GPU video memory, where this information is natively being generated \cite{Richter_2016_ECCV}. In a modern game engine, such as UE, this information can be exposed through post-processing materials and rendered and exported using render targets or through the APIs of the available UE-based open-source simulators \cite{dosovitskiy2017carlaopenurbandriving}. On the other hand, real-world datasets that depict scenes similar to those in the game can be found in public benchmarks employed in deep learning research or captured manually using cameras in real-world environments. Examples include urban driving datasets such as Cityscapes \cite{cityscapes} and KITTI \cite{kitti} that have been employed to enhance the photorealism in simulated automotive scenarios or urban scenes in games and simulators, such as GTA V \cite{Richter_2021} and CARLA \cite{pasios2024carla2realtoolreducingsim2real}.

\subsection{Robust Unpaired Image-to-Image Translation Network}

Having access to the appropriate in-game and real-world datasets enables the training of a robust unpaired Im2Im translation method to produce a photorealism-enhanced version of the game dataset. In this work, we utilize the Enhancing Photorealism Enhancement (EPE) framework proposed by Richter et al. \cite{Richter_2021}, which to date has illustrated SotA performance in enhancing the photorealism of computer-graphics-based applications and is the only method that was presented as an approach specifically designed for photorealism enhancement in games. In detail, EPE is a generative adversarial network that consists of a generator that enhances the photorealism of synthetic images and a discriminator that tries to classify whether a frame originates from the generator or the real-world dataset. Since the model is trained in an adversarial manner and the discriminator may learn to distinguish the images by distribution differences between the synthetic and the real images, the method employs a patch-matching approach prior to training to match patches between the source (game) and the target (real-world) domain that depict similar objects. This is performed by extracting features from these patches through a VGG-16 architecture and executing a similarity search by utilizing the Facebook AI Similarity Search (FAISS) library \cite{faiss}. Additionally, the generator employs the G-Buffers generated by the game engine and processes them through a G-Buffer encoder, which consists of multiple streams for each semantic class of the game as defined by the stencil. This enables the generator to treat individual virtual objects differently and utilize only the G-Buffers that are informative for each semantic class. For instance, the sky in a virtual environment does not contain information about the material properties, and thus, the respective G-Buffers (e.g., metallic, base color, and roughness) always contain black pixels in that region. Moreover, considering that most of the G-Buffers are independent of the appearance and the events that occur in the rendered scene (e.g., the base color of an object remains the same despite the view or the illumination changes), they enable the generator to produce temporally consistent results. As a result, EPE has been successfully employed on a synthetic dataset designed for video classification (crowd abnormal behavior detection) \cite{pscsi}, leading to an increase in the classification performance on cross-dataset evaluation on real-world data. Specifically, in the domain of crowd abnormal behavior, temporal inconsistencies would result in false positives of crowd behavior anomalies, leading to a negative impact on the accuracy. 

Finally, the discriminator employs a robust semantic segmentation network, MSEG \cite{mseg}, that was trained on seven different semantic segmentation datasets (e.g., Cityscapes) with a unified semantic taxonomy of 194 classes and is inferred on both the real-world and the synthetic images used during training. This enables the guidance of the discriminator without requiring annotations for the real-world dataset or remapping to enable compatibility with the ones generated by the stencil of the game.

\subsection{Paired Image-to-Image Translation Network}

After robustly generating a photorealism-enhanced dataset based on a non-real-time unpaired Im2Im translation that is semantically consistent with the video game dataset, a paired Im2Im translation method can be employed and trained on the in-game and photorealism-enhanced pairs. In this work, Pix2PixHD \cite{wang2018pix2pixHD} was selected since it's one of the fastest methods in terms of inference speed at high-resolution images, which are important for video games. While smaller models with increased speed compared to Pix2PixHD, such as AsapNet  \cite{asapnet}, also exist, these are limited to specific tasks (i.e., label to image) that are not applicable for photorealism enhancement. Moreover, Pix2PixHD is specialized for paired image-to-image translation, which generally produces more accurate and visually coherent results than models that support both paired and unpaired translation \cite{CUT, munit}. In detail, Pix2PixHD builds upon the original Pix2Pix model \cite{pixpixold} by adding a coarse-to-fine generator network, multi-scale discriminators, and feature matching losses. These additions enable it to generate visually coherent and detailed results even at resolutions of 2048×1024 with an inference delay of around 20-30 ms (33.33-50 FPS) with a GTX 1080Ti GPU. One benefit of this method is that it supports supervised training with ground truth pairs, allowing for the use of pixel-level losses (i.e., L1, perceptual loss, and adversarial loss) to direct the translation. This generally leads to more stable and accurate outputs than unpaired Im2Im translation techniques, particularly when fine visual details and consistency with game geometry matter.

\subsection{Game Integration}

The need for integrating deep-learning models has emerged into various frameworks that enable seamless deployment across platforms such as mobile devices, edge computing units, and cloud infrastructure. Among them, Open Neural Network Exchange (ONNX) has become an essential framework for deploying machine learning based models into products due to its flexibility, which is ideal for real-time inference scenarios. Combined with ONNX Runtime\footnote{https://onnxruntime.ai/}, optimized inference speed can be achieved by hardware accelerators such as TensorRT\footnote{https://developer.nvidia.com/tensorrt}, which enables the inference in mixed precision utilizing the tensor cores that are available in modern gaming and AI GPUs. Due to these significant features of ONNX Runtime, the library was also implemented in the latest version of UE, Unreal Engine 5 (UE5) (UE5)\footnote{https://github.com/microsoft/OnnxRuntime-UnrealEngine}, and Unity\footnote{https://github.com/asus4/onnxruntime-unity}, enabling the utilization of generative models as post-process filters. Since the proposed framework removes the requirement of additional G-Buffers imposed by robust unpaired Im2Im translation methods, the approach is fully compatible with the UE5 and Unity ONNX Runtime libraries, which are built solely for inference on the final rendered frame.

\section{ Experiments} \label{section_experiments}

In this section, we compare EPE with the proposed framework in terms of visual quality, generalization capacity, semantic preservation, and inference speed. In addition, we investigate the temporal consistency of the proposed framework.

\subsection{Datasets}

\textbf{CARLA2Real-UE4} \cite{pasios2024carla2realtoolreducingsim2real} is a dataset extracted from the CARLA simulator in order to serve as training input to EPE. The dataset includes $20,014$ training frames, each accompanied by the Unreal Engine 4 (UE4) G-buffers used for rendering, semantic segmentation annotations for 28 classes, and the corresponding EPE results, translated towards the characteristics of the  Cityscapes and KITTI real-world datasets. The dataset also provides $1,000$ test images for the model targeting the KITTI characteristics and $10,792$ test images for the Cityscapes variant. Moreover, the CARLA2Real dataset is extended with $500$ additional frames generated using the newer UE5 version of the CARLA simulator, \textbf{CARLA2Real-UE5} \cite{PasiosNikolaidis2025_EUSIPCO}, which are similarly translated towards KITTI and Cityscapes using the EPE models trained on the UE4 frames. \textbf{CrowdFlow} \cite{8639113} is a dataset designed to benchmark optical flow methods. Specifically, it was generated using synthetic environments rendered in UE4 and contains 10 videos with a total of 3,600 frames. Five videos were recorded from static cameras, and the remaining five from dynamic, drone-mounted ones. Each frame includes ground-truth annotations for optical flow fields, person trajectories, and dense pixel trajectories.

\textbf{Cityscapes} \cite{cityscapes} is a real-world dataset that is considered a standard benchmark for photorealism enhancement via Im2Im translation. Cityscapes images were captured across 50 cities in Germany in good weather conditions. The dataset includes pixel-level semantic segmentation annotations for 30 unique categories. The original dataset contains $5,000$ images, but it was also extended to a larger version with $20,000$ frames accompanied by coarse annotations. \textbf{KITTI} \cite{kitti} is a real-world dataset widely used in autonomous driving research. It was captured in the city of Karlsruhe, Germany, and provides $15,000$ frames accompanied by information extracted from laser scans, high-precision GPS measurements, and IMU accelerations. The dataset also contains $200$ images that are annotated at the pixel level for semantic segmentation tasks, offering 34 unique semantic classes.

\subsection{Metrics}

CLIP Maximum Mean Discrepancy (CMMD) \cite{cmmd} is a recent metric for evaluating generative models based on CLIP’s robust, semantically rich feature space. CMMD computes the Maximum Mean Discrepancy using a characteristic Gaussian RBF kernel, avoiding assumptions that embeddings follow a Gaussian distribution. By operating directly in CLIP space, it captures higher-level perceptual attributes and aligns more closely with human judgments. Empirically, CMMD was shown to correlate better with human perception of visual appeal than metrics such as FID \cite{fid}. Based on the importance of user likeness and satisfaction in video games, we consider this metric ideal for our evaluation, and thus, it was employed in the experiments.

In addition to CMMD, we use a metric for evaluating the semantic preservation capabilities of the framework. In detail, the mean Intersection over Union (mIoU) is employed, which measures the overlap between predicted and ground-truth semantic segmentation masks by averaging intersection-over-union across classes. Finally, for the optical flow and temporal consistency investigation of the proposed framework, we employ the Endpoint Error, which measures the Euclidean distance between predicted and ground-truth flow vectors at each pixel.

\subsection{Setup \& Implementation Details}

To train REGEN, we employed the CARLA2Real-UE4 dataset and specifically the same 15,011 frames used in \cite{pasios2024carla2realtoolreducingsim2real} to train EPE. These frames were used to construct paired datasets between the raw CARLA renderings and their photorealism-enhanced counterparts, enabling supervised training of Pix2PixHD at a resolution of 960×540 (the resolution of the CARLA2Real-UE4 dataset). In detail, Pix2PixHD was trained for a total of 9 epochs for two trained models: first, a Cityscapes-targeted model was trained using pairs of rendered CARLA frames and their Cityscapes-enhanced counterparts from \cite{pasios2024carla2realtoolreducingsim2real}; and second, a KITTI-targeted model was trained using the same rendered frames paired with KITTI-enhanced images. After training, REGEN was applied to the $10,792$ Cityscapes-targeted test images and the $1,000$ KITTI test images, and its outputs were evaluated using CMMD, measuring similarity against the respective real-world datasets (i.e., the $5,000$ Cityscapes and the $15,000$ KITTI  images). The same CMMD evaluation was repeated for the initial EPE-generated test images included in CARLA2Real-UE4 \cite{pasios2024carla2realtoolreducingsim2real}, allowing a direct comparison between REGEN and the initial robust photorealism enhancement method. To further assess generalization capabilities, the same evaluation protocol was applied to the additional set of $500$ frames of the CARLA2Real-UE5 dataset \cite{PasiosNikolaidis2025_EUSIPCO}, which was rendered from the newer UE5 version of the CARLA simulator. Furthermore, REGEN was cross-evaluated on the CrowdFlow dataset (for both Cityscapes and KITTI target domains) to examine its ability to improve photorealism in completely unseen environments. Since CrowdFlow does not provide the required G-Buffer information, EPE could not be applied to this dataset.

To assess semantic preservation (whether REGEN maintains scene content without introducing unrealistic artifacts), an additional experiment was conducted using semantic segmentation models. The rationale behind this evaluation is that segmentation accuracy on the data generated by REGEN and EPE should remain similar if photorealistic enhancement does not alter the semantics of the scene. Since the ground-truth labels remain unchanged after the translation of the images, unrealistic objects or alterations to the semantics introduced by REGEN will result in a decrease in semantic segmentation accuracy compared to the one achieved with the data produced by EPE. For Cityscapes, a pretrained DeepLabV3 \cite{Chen2017RethinkingAC} model trained on the 19-class Cityscapes subset was used. For KITTI, the same model was further fine-tuned on the $200$ KITTI images, randomly split into 80\% training and 20\% testing sets. Finetuning was performed for 5 epochs, after which overfitting began to occur due to the limited dataset size. These models were then applied to the images produced by both REGEN and EPE for the test frames of CARLA2Real-UE4 and the entire CARLAReal-UE5 dataset. CrowdFlow was excluded from this experiment since it does not provide semantic segmentation annotations.

Temporal consistency was evaluated using the CrowdFlow dataset, the only one in our study that is provided in video form and is designed for optical flow benchmarking. For each sequence, the REGEN models (trained to translate CARLA2Real-UE4 towards the characteristics of Cityscapes and KITTI) were applied to produce enhanced results of the frames. Optical flow fields were then computed on both the original rendered frames and the enhanced frames using a traditional optical flow estimation algorithm, namely the Gunnar-Farneback optical flow algorithm \cite{Farneb}, and compared against ground-truth flow. The experiment was performed on both the background-only pixels and the ones that depicted the crowd with dynamic movements, similarly to \cite{8639113}. Since video games involve frequent and intense movement, we chose to use in this experiment only the videos captured from the dynamic (moving) camera, as they provide a more challenging scenario that can lead im2im translation to more temporal artifacts (e.g., flickering or semantic inconsistencies) in comparison with images from static cameras.  Preservation of optical-flow error—i.e., minimal deviation between the initial synthetic rendered and the enhanced sequences- indicates that REGEN maintains temporal coherence and does not introduce frame-to-frame inconsistencies.

All experiments were conducted on a gaming system equipped with an Intel i7-14700F CPU, an NVIDIA RTX 4090 GPU with 24 GB of VRAM, and 64 GB of DDR4 system memory. This setup was also used to benchmark the inference speed of both the proposed approach and EPE. Specifically, for benchmarking, we used the modified EPE architecture from \cite{pasios2024carla2realtoolreducingsim2real}, which achieves higher inference speed compared to the original implementation in \cite{Richter_2021}. Both EPE and REGEN were executed within the CARLA simulator in synchronous mode (i.e., the simulator waits for the models before rendering the next frame) for 100 simulation steps to evaluate whether the resulting frame rate remained consistent.

Across the described experimental setup, the same images were used between the methods (i.e., the same rendered images and the results of EPE and REGEN) to avoid potential bias in the comparisons. In addition, downscaling was applied to the images only prior to the inference of the DeepLabV3 models to a resolution of 720x360. Finally, for within-dataset evaluations (i.e., CARLA2Real-UE4), only the test sets of the datasets were employed. On the other hand, for cross-dataset experiments (i.e., CARLA2Real-UE5 and CrowdFlow), the entire datasets were used.

\subsection{Results \& Discussion}

In this section, we present and discuss the results of the REGEN framework in comparison to EPE. We first evaluate visual similarity to real-world datasets and benchmark inference speed, followed by an analysis of semantic preservation using pretrained segmentation models applied to the output of EPE and REGEN. Finally, we examine temporal consistency in video sequences to assess whether REGEN maintains frame-to-frame (temporal) coherence.

\begin{table}[h!]
\centering
\caption{Comparison of visual similarity between engine-generated data, the EPE and REGEN outputs, and the corresponding real-world benchmarks (Cityscapes and KITTI) based on the CMMD metric (lower is better). Both EPE and REGEN are trained on the CARLA2Real-UE4 dataset.}
\resizebox{\columnwidth}{!}{
\begin{tabular}{lcc|cc|cc}
\hline
Method & \multicolumn{2}{c|}{CARLA2Real-UE4} & \multicolumn{2}{c|}{CARLA2Real-UE5} & \multicolumn{2}{c}{CrowdFlow} \\
 & Cityscapes & KITTI & Cityscapes & KITTI & Cityscapes & KITTI \\
\hline
UE        & 4.732 & 5.269 & 4.734 & 4.790 & 6.653 & 6.318 \\
%EPE/CUT  & 3.577 & 3.777 & 3.525 & 3.431 & -     & -     \\
EPE       & 3.420 & 4.233 & 3.764 & 3.197 & -     & -     \\
REGEN     & 3.232 & 4.062 & 3.483 & 3.170 & 5.379 & 4.643 \\
\hline
\end{tabular}
}
\label{tab:cmmd_comparisons}
\end{table}

\subsubsection{Visual Similarity \& Runtime Performance}

\begin{figure}[htbp]
    \hspace*{0mm} % optional fine control
    \includegraphics[width=0.47\textwidth]{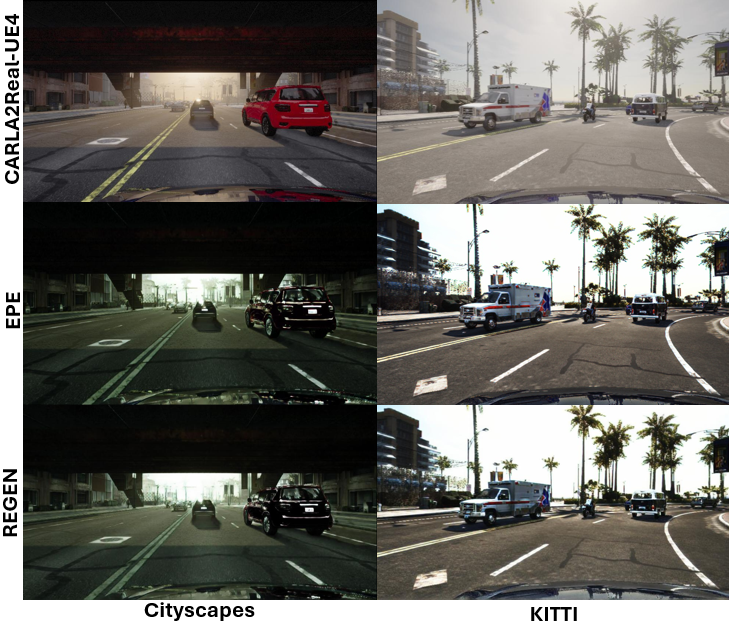}
    \caption{Visual comparison of the images generated by EPE and REGEN towards the characteristics of Cityscapes and KITTI, when given as input a CARLA2Real-UE4 frame from the test set.}
    \label{fig:metho_visual_comp}
\end{figure}

\begin{figure}[htbp]
    \centering
    \includegraphics[width=0.48\textwidth]{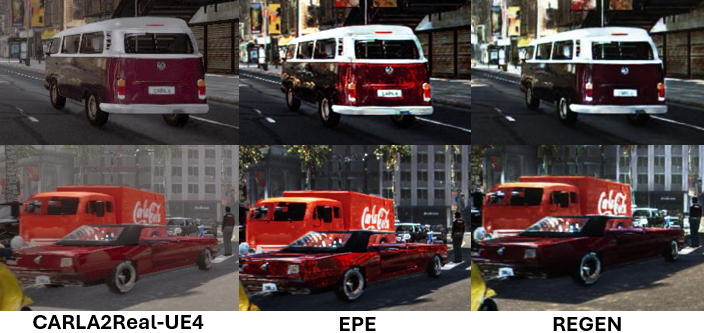}
    \caption{Example failure cases of EPE, where unrealistic glossiness and material artifacts appear on vehicle surfaces compared to the initial render and the results produced by the proposed method.}
    \label{fig:epe_errors}
\end{figure}

In this subsection, we present and discuss the results of the REGEN framework compared to the initial outputs of EPE for CARLA2Real-UE4 translated toward the characteristics of both Cityscapes and KITTI. Additionally, the models are evaluated on the newer version of the CARLA simulator (CARLA2Real-UE5), which was unseen during training for both methods. Finally, REGEN, as a method that does not rely on additional information extracted from the game engine, was also cross-evaluated on CrowdFlow. As shown in Table \ref{tab:cmmd_comparisons}, for both CARLA2Real-UE4 and CARLA2Real-UE5 datasets, REGEN achieves slightly improved CMMD scores, indicating that its outputs are more visually appealing and closer to the target datasets (Cityscapes and KITTI, respectively). In detail, according to \cite{pasios2024carla2realtoolreducingsim2real}, EPE is prone to some failure cases, such as hallucinations of palm trees (this failure case was not reported for the translation results on the CARLA simulator) and unrealistic glossiness on the vehicles, which can lead to an increase in CMMD. As illustrated in Fig. \ref{fig:metho_visual_comp} and Fig. \ref{fig:epe_errors}, REGEN does not learn these unrealistic artifacts. Therefore, we attribute the small improvement observed for REGEN compared to EPE to the aforementioned failure cases of the latter.

Concerning model runtime performance, the inference time and FPS are reported in Table \ref{tab:performance_4090_selected} as mean and standard deviation over 100 simulation steps to evaluate stability.  The GPU memory utilization (measured for the entire system, including both the models and the CARLA simulator) is also reported. Both models (REGEN and EPE) were evaluated at resolutions of 960×540 (as in \cite{Richter_2021}), 1280×720 (720p), and 1920×1080 (1080p). It is evident that while EPE has slightly lower GPU memory requirements, it cannot run in real-time, achieving only 2.58 FPS at 960×540. In contrast, REGEN maintains a balance between speed and memory, reaching $32$ FPS at 960×540, $25,57$ FPS at 720p, and $11$ FPS at 1080p. Since no quantization or optimization (e.g., TensorRT) was applied, FPS could be further improved. Moreover, real-time inference at lower resolutions enables the potential of combining REGEN with upscaling techniques, such as NVIDIA’s Deep Learning Super Sampling (DLSS), to further improve performance.

\subsubsection{Semantic Preservation}

We also compared REGEN and EPE in terms of their ability to preserve the semantic structure of the original rendered images. Specifically,  DeepLabV3 semantic segmentation models pretrained on real-world datasets are applied to the generated frames of each method. As shown in Table \ref{tab:all_methods_cityscapes}, for CARLA2Real-UE4, REGEN preserves the semantic structure in a way similar to the initial robust Im2Im translation method (EPE). For Cityscapes-enhanced frames, the mIoU is nearly identical, and for KITTI, only a small decrease is observed. However, this decrease would be significantly larger for unpaired image-to-image translation methods due to the distribution difference between the rendered images and the real-world (see Appendix \ref{appendix_baselines}). The same observation is evident for CARLA2Real-UE5 frames, which were unseen during training for both methods.

\subsubsection{Temporal Consistency}

\begin{figure}[htbp]
    \centering
    \includegraphics[width=0.5\textwidth]{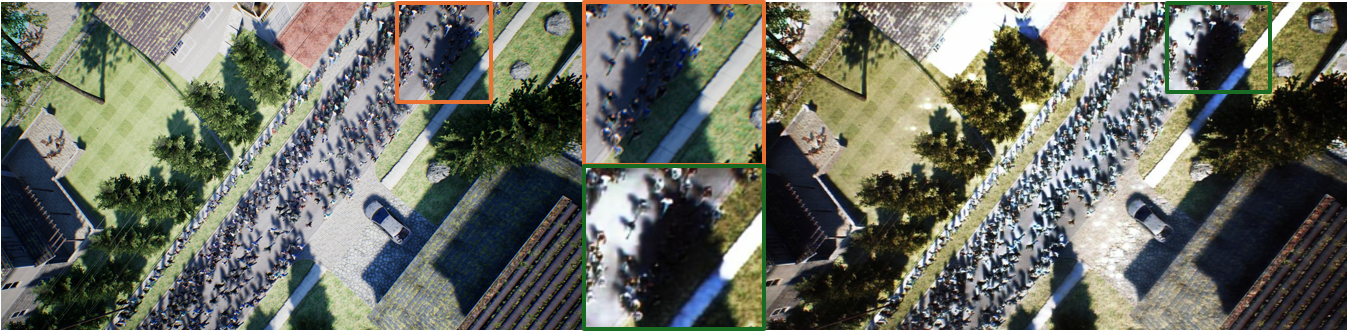}
    \caption{Translation result of REGEN on a CrowdFlow frame (left) towards the characteristics of KITTI (right).}
    \label{fig:crowd_flow_appendix}
\end{figure}

In this subsection, we investigate the temporal consistency of the proposed method. REGEN was applied to videos from the CrowdFlow dataset for both Cityscapes and KITTI characteristics. Optical flow was extracted using the Gunnar-Farneback algorithm for the synthetic rendered videos as well as for the REGEN–Cityscapes and REGEN–KITTI variations. The resulting optical flow was then compared to the ground-truth flow. As shown in Table \ref{tab:optical}, the optical flow error at background pixels (which do not include crowd) is not only maintained but slightly reduced by REGEN. This improvement may stem from the smoothing of artifacts present in the rendered scenes, such as those introduced by anti-aliasing. Conversely, for pixels related to the crowd, a small increase in error is observed. Since the method significantly increases shadow intensity, which in some instances leads to full or partial occlusions of the crowd, as shown in Fig. \ref{fig:crowd_flow_appendix}, this can lead to wrong optical flow estimation compared to the ground truth and therefore justifies this small increase in Endpoint Error.

\begin{figure*}[htbp]
    \centering
    \includegraphics[width=1\textwidth]{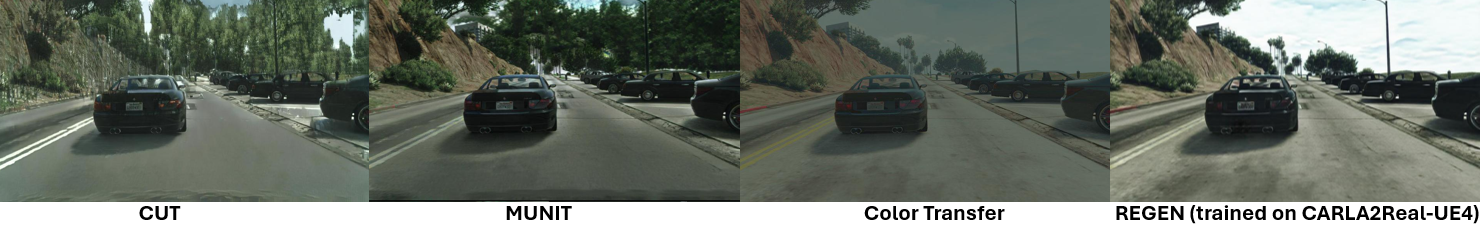}
    \caption{Visual comparison of the images generated by CUT, MUNIT, Color Transfer, and REGEN, when given as input a PFD frame from the test set.}
    \label{fig:baselines}
\end{figure*}

\begin{table*}[h!]
\centering
\caption{Runtime performance (FPS, latency) and VRAM usage of EPE and REGEN on an RTX 4090 at selected resolutions.}
\begin{tabular}{lccc|ccc|ccc}
\hline
Method & \multicolumn{3}{c|}{960x540} & \multicolumn{3}{c|}{1280x720 (720p)} & \multicolumn{3}{c}{1920x1080 (1080p)} \\
 & MS $\downarrow$ & FPS $\uparrow$ & VRAM $\downarrow$ & MS $\downarrow$ & FPS $\uparrow$ & VRAM $\downarrow$ & MS $\downarrow$ & FPS $\uparrow$ & VRAM $\downarrow$ \\
\hline
EPE & 387.98 ± 1.89 & 2.58 ± 0.01 & 8.5GB & 705.85 ± 5.51 & 1.42 ± 0.01 & 9.6GB & 1801.30 ± 6.18 & 0.56 ± 0.00 & 12.9GB \\
REGEN & 32.13 ± 0.66 & 31.12 ± 0.64 & 9GB & 39.11 ± 0.59 & 25.57 ± 0.39 & 10.3GB & 91.03 ± 1.45 & 10.99 ± 0.18 & 14GB \\
\hline
\end{tabular}
\label{tab:performance_4090_selected}
\end{table*}

\begin{table}[h!]
\tiny
\centering
\caption{Comparison of semantic segmentation accuracy on the translation results of EPE and REGEN (both trained on CARLA2Real-UE4) using mIoU.}
\resizebox{\columnwidth}{!}{
\begin{tabular}{lcc|cc}
\hline
Method & \multicolumn{2}{c|}{CARLA2Real-UE4} & \multicolumn{2}{c}{CARLA2Real-UE5} \\
 & Cityscapes & KITTI & Cityscapes & KITTI  \\
\hline
EPE       & 24.76\% & 18.36\% & 20.19\% & 19.53\% \\
REGEN     & 24.66\% & 17.82\% & 20.87\% & 19.29\% \\
\hline
\end{tabular}
}
\label{tab:all_methods_cityscapes}
\end{table}

\begin{table}[h!]
\centering
\caption{Optical flow error of the initial rendered CrowdFlow dataset, the KITTI, and Cityscapes variations produced by REGEN (trained on CARLA2Real-UE4) compared to the ground truth flows. The Endpoint Error metric is employed to calculate error only for the background and person-depicting pixels.}
\scalebox{0.95}{
\begin{tabular}{lcc}
\hline
Method & \multicolumn{2}{c}{CrowdFlow} \\
 & Background & Crowd   \\
\hline
UE                    & 2.491 & 0.996 \\
REGEN-Cityscapes      & 2.400 & 1.056 \\
REGEN-KITTI           & 2.408 & 1.083 \\
\hline
\end{tabular}
}
\label{tab:optical}
\end{table}

\section{Conclusion} \label{section_conclusion}

In this paper, a two-stage framework for real-time photorealism enhancement in games is proposed. We developed a pipeline that learns to enhance the photorealism of UE environments, towards the  Cityscapes and KITTI real-world datasets, by employing a robust unpaired Im2Im translation method (EPE) to transform the problem into an easier, paired Im2Im translation task that can be performed with lightweight paired Im2Im architectures such as Pix2PixHD. To evaluate the contribution of the proposed framework, we employed both the photorealism-enhanced results of EPE and the ones produced by the introduced framework. The results illustrated that our approach can slightly improve the visual realism of EPE in both within- and cross-dataset evaluations and increase the frames per second up to 12 times. In addition, it was illustrated that it can adhere to the semantic preservation of EPE and maintain temporal consistency similar to the initial rendered sequences.

This short paper outlines promising directions for future research on paired Im2Im translation methods specifically designed and optimized to accurately replicate the characteristics of generated game frames produced by the current robust Im2Im translation approaches for real-time photorealism enhancement, similarly to methods specifically optimized in terms of inference for other domain-specific tasks, such as labels to image \cite{asapnet}.

\bibliographystyle{IEEEtran}
\bibliography{bibliography}

%\newpage
\appendix

\subsection{Comparisons with Lightweight Unpaired Image-to-Image Translation Methods} \label{appendix_baselines}

In this appendix, we compare REGEN with lightweight unpaired Im2Im translation methods, namely, CUT, MUNIT \cite{munit}, and Color Transfer \cite{946629} to illustrate that the introduced framework, in comparison to these methods, even in a cross-dataset setting, can still maintain a higher level of semantic consistency. In detail, in \cite{Richter_2021}, the authors have trained on the Playing for Data (PFD) \cite{Richter_2016_ECCV} ($12,403$ training, $6,382$ validation, and $6,181$ test images) three baselines, CUT, MUNIT, and color transfer. The resulting images are provided publicly for $4,204$ frames of the PFD test set towards the characteristics of Cityscapes (results for the KITTI dataset are not published). Therefore, we applied REGEN trained on CARLA2Real-UE4 towards the characteristics of the same dataset on the same frames, and applied the DeepLabV3 model trained on Cityscapes on the resulting images of all methods (PFD is provided with semantic segmentation annotations compatible with the Cityscapes annotation scheme). The EPE model used in the experiments cannot be employed on GTA due to G-Buffer incompatibility with UE. The results are illustrated in Table \ref{tab:baselines_table}, where it is evident that  REGEN achieves the highest mIoU across all the baseline unpaired methods, despite not being explicitly trained on the PFD dataset. In contrast, CUT and MUNIT introduce visual artifacts, such as vegetation in the sky, that confuse the semantic segmentation model and lead to misclassifications, reducing accuracy and limiting their applicability in video games. On the other hand, Color Transfers fails to effectively capture the characteristics of the Cityscapes dataset. These observations are visually depicted in Fig. \ref{fig:baselines}.

\begin{table}[h]
\caption{Comparison of semantic segmentation accuracy on the REGEN and the baseline unpaired im2im translation method outputs on the PDF dataset towards Cityscapes using mIoU. REGEN is trained on the CARLA2Real-UE4 dataset.}
\centering
\begin{tabular}{lccccc}
\hline
 Dataset & CUT & MUNIT & COLOR TRANSFER & REGEN \\
\hline
PFD \cite{Richter_2016_ECCV} & 16.44\% & 19.19\% & 19.76\% & 22.23\% \\
\hline
\end{tabular}
\label{tab:baselines_table}
\end{table}

\end{document}